\documentclass[runningheads]{llncs}
\usepackage[T1]{fontenc}
\usepackage{graphicx}

\usepackage{adjustbox}

\usepackage[utf8]{inputenc}

\usepackage{cite}
\usepackage{amsmath,amssymb,amsfonts}
\usepackage{algorithmicx}

\usepackage{algcompatible}
\usepackage{algpseudocode}

\usepackage{algcompatible}
\usepackage{tcolorbox}
\usepackage{graphicx}
\usepackage{textcomp}
\usepackage{xcolor}
\usepackage{cuted}
\usepackage{mathtools}
\usepackage{cuted}
\usepackage{mathabx}
\usepackage{enumitem}
\usepackage{siunitx}
\usepackage{placeins}

\edef\oldtt{\ttdefault}
\usepackage[scaled]{beramono}
\usepackage[T1]{fontenc}
\renewcommand{\ttdefault}{\oldtt}
\newcommand{\bera}[1]{{\fontfamily{fvm}\selectfont #1}}

\usepackage{inconsolata}
\usepackage[T1]{fontenc}

\def\BibTeX{{\rm B\kern-.05em{\sc i\kern-.025em b}\kern-.08em
    T\kern-.1667em\lower.7ex\hbox{E}\kern-.125emX}}

\usepackage{array,multirow,multicol}
\usepackage{booktabs}
\usepackage[caption=false,font=normalsize,labelfont=sf,textfont=sf]{subfig}
\usepackage{stfloats}
\usepackage{url}
\usepackage{verbatim}
\usepackage{longtable}
\usepackage{pifont}
\usepackage{wrapfig}
\usepackage{mathrsfs}
\usepackage[font=normalsize, labelfont=bf]{caption}
\usepackage{subcaption}
\usepackage{hyperref}

\RequirePackage{algorithm}

\usepackage{float}





\newcommand{\nop}[1]{}

\newcommand{\system}{RecLLMs}
\usepackage[svgnames]{xcolor}

\usepackage{datetime}

\newcommand{\R}{\mathbb{R}}
\newcommand{\cN}{\mathcal{N}}

\usepackage{subcaption}
\usepackage{wrapfig}
\begin{document}
\title{Lightweight Fairness for LLM-Based Recommendations via Kernelized Projection and Gated Adapters}

\author{Nan Cui \and
Wendy Hui Wang\and
Yue Ning}
\authorrunning{N.\ Cui et al.}
\institute{Department of Computer Science, Stevens Institute of Technology, Hoboken NJ, USA\\
\email{ncui@stevens.edu, hwang4@stevens.edu, yue.ning@stevens.edu}}

\maketitle             
\thispagestyle{plain}
\pagestyle{plain}
\begin{abstract}
Large Language Models (LLMs) have introduced new capabilities to recommender systems, enabling dynamic, context-aware, and conversational recommendations. However, LLM-based recommender systems inherit and may amplify social biases embedded in their pre-training data, especially when demographic cues are present. Existing fairness solutions either require extra parameters fine-tuning, or suffer from optimization instability. We propose a lightweight and scalable bias mitigation method that combines a kernelized Iterative Null-space Projection (INLP) with a gated Mixture-of-Experts (MoE) adapter. Our approach estimates a closed-form projection that removes single or multiple sensitive attributes from LLM representations with no additional trainable parameters. To preserve task utility, we introduce a two-level MoE adapter that selectively restores useful signals without reintroducing bias. Experiments on two public datasets show that our method reduces attribute leakage across multiple protected variables while maintaining competitive recommendation accuracy.

\end{abstract}

\section{Introduction}\label{sec:intro}

The advent of Large Language Models (LLMs) is reshaping the design and capabilities of recommender systems. By leveraging the rich semantic understanding and generative abilities of LLMs, new systems can move beyond static suggestions to support dynamic, context-aware, and interactive recommendations. LLM-based recommender systems (\system)~\cite{wu2024survey} enable users to engage in natural conversations while receiving personalized and situationally relevant suggestions. This shift opens up new possibilities for personalization, including handling complex user intents, understanding implicit preferences, and adapting to evolving user contexts in real time. While prompt-only methods using general-purpose LLMs may be sufficient for small-scale or low-stakes applications, they often fall short in accuracy, particularly in tasks like named product generation. These limitations point to the need for tighter integration between LLMs and collaborative filtering, user profiles, and retrieval components to build scalable and reliable recommendation systems.
Prior work such as Liao~\textit{et al.}~\cite{liao2024llara} jointly aligns embeddings from traditional recommender systems~\cite{ZhangFZBWH25,NgoN24} with LLM embeddings and produce more accurate recommendations.

Despite the recent success of RecLLMs, they inevitably inherit - and sometimes amplify - social biases present in their pre‑training corpora \cite{GallegosRBTKDYZA24}. When sensitive information such as \emph{gender}, \emph{age}, or \emph{occupation} appear in a prompt, the generated recommendation list  can privilege or penalize specific demographic groups. Existing bias mitigation techniques employ two main strategies. The first calibrates \emph{item exposure}  \cite{JiangBZW0F024}, which balances item popularity but ignores user‐side disparities. The second protects the \emph{user-side fairess} with gradient‐reversal discriminators (notably UP5~\cite{HuaGXJLZ24}). While effective, these adversarial modules introduce large scale of trainable parameters and remain brittle to optimization hyper‑parameters. For example, UP5 ~\cite{HuaGXJLZ24} mitigates the per-attribute retraining by adding an attention controller, yet the adversarial coupling still destabilizes optimization. There are also bias diagnostic tools that offer purely diagnostic probes, such as CFaiRLLM~\cite{cfairllm2024} construct a neutral prompt and an otherwise identical prompt that names one (or an intersection of) sensitive attributes (e.g., gender, age, etc.), then compares the two top-$k$
 lists with their specialized metrics; it flags discrepancies as consumer-side unfairness but intentionally proposes no debiasing remedy. 
 In practice, an LLM primarily serves its utility task, and any debiasing step that requires additional training cycles or add-on networks will lead to increased computational costs.
 At present, no existing method focuses on optimization-effective debiasing for \system~from an user-side view.
Thus, we need a ``\emph{lightweight}'' alternative that removes sensitive information without adding optimization headaches. 

Iterative Null‑space Projection (INLP)~\cite{RavfogelEGTG20} answers this call because it debiases by \emph{closed‑form linear algebra} rather than by re‑optimizing the whole network. Given a well-trained RecLLM, the recipe is simple: 
(i) \emph{freeze} the language model,  
(ii) \emph{extract} the sequence-level representation (i.e., embedding vector for an input prompt) from the model, train a 
\emph{probe} (i.e., a linear classifier) on this vector to predict a sensitive label (say, gender), and 
(iii) project the sequence-leval representation onto the \emph{null subspace}, the set of directions orthogonal to the probe’s weight vector. In plain language, the probe is a measure for bias, and the null subspace is the part of representation space it cannot see. We get the projection matrix from a least‑square solver, so no gradient ever touches the backbone.
After applying the projection matrix to each embedding vector, this projected representation replaces the original vector in the language model before its output layer. Because of this algebraic nature, INLP has become a popular off‑the‑shelf debiasing tool for word embeddings~\cite{GonenG19} and BERT encoders~\cite{GarimellaAKYNCS21}. 

However, two obstacles emerge in practice. First, a single linear probe cannot capture the non‑linear signals an LLM exploits, leaving left-over bias. Second, mitigating the bias of multiple sensitive attributes would require a separate probe - and a separate projection - for each attribute. 
Stacking those projections can erase task‑useful variance and will rapidly reduce the performance of the model.

To tackle the first obstacle (i.e., \emph{non‑linear leakage}), we {kernelize the probe with Random Fourier Features (RFF)~\cite{RahimiR07}, thereby upgrading linear INLP to a universal approximation while still yielding a \emph{closed‑form} projector. 
We further inject isotropic Gaussian perturbation into each hidden state during probe fitting. This operation widens the decision margin and makes the null space robust to the representation shift introduced by fine-tuning.
The resulting matrix is stored as a non‑trainable buffer with no gradients, no optimizer state, and negligible overhead, which we named ``lightweight''. 

To address the second obstacle (i.e., \emph{handling multiple sensitive attributes}), we attach $\mathbf P$ with a two‑level gated Mixture of Experts (MoE) adapter. 
MoE adapter consists of two gates. 
The first gate reads the current representation and outputs a probability vector that softly blends a set of attribute-specific projectors. This allows the model to control the intensity of removing each attribute bias. 
The partly purified embedding then enters several low-rank LoRA experts; 
inside every expert a second residual gate scales its own update before all updates are summed into a single task vector, 
which is then lightly fine-tuned after debiasing to improve main-task accuracy. This ``erase-then-repair'' pipeline lets \system\ provide fair and accurate recommendations.

Our work has two main contributions: (1) we propose a \emph{kernelized INLP projector}  that removes the information of the sensitive attributes in one closed-form step, incurring zero optimization cost; (2) we devise a \emph{two-level gated MoE adapter} that recovers task utility in the multi-attribute setting while introducing only $\mathcal{O}(k)$ additional parameters for $k$ sensitive attributes.
We validate the approach on two public LLM-based recommendation benchmarks, demonstrating consistent fairness improvements without sacrificing recommendation quality.

\section{Related Work}\label{sec:related}

\textbf{LLM-Based Recommendation Systems}
Essentially, the research on RecLLMs can be categorized into two types: (i) {\em Prompt-only pipelines}, which translate recommendation queries as free-text questions that are answerable by general-purpose LLMs. Gao \textit{et al.}~\cite{Gao2023ChatRec} retrieve review snippets to ground ChatGPT responses. Geng \textit{et al.}~\cite{Geng0FGZ22} formulate the recommendation tasks in a unified text-to-text fashion. 
(ii) {\em Parameter-efficient fine-tuning based methods} which equip the instruction-tuned backbones with lightweight adapters. Ngo \textit{et al.}~\cite{NgoN24} design the recommender systems based on instruction-tuned LLMs, and demonstrates that these systems outperform the conventional collaborative-filtering based recommender systems \cite{Fan2023LLMs4Rec}. Some recent works utilize a hybrid architecture that injects entity embeddings 
into instruction-tuned backbones ~\cite{liao2024llara}. Unlike prompt-only systems, our work fine-tunes an open Llama model~\cite{abs-2407-21783} due to its full exploit of identifier-level interaction signals 
preservation of the conversational flexibility.


\textbf{Fairness in RecLLMs}
Most of the research of fairness in RecLLMs focus on mitigating the bias at either the user side or the item side.  
Early work re-ranks or re-weights predicted preference scores
to curb popularity bias and protect under-served items ~\cite{Burke2022MASCFairRec}, whereas Ekstrand \textit{et al.}~\cite{EkstrandBD19} regularize latent factors and Wu \textit{et al.}~\cite{Wu2022PFRec} employ gradient-reversal debiasing to improve user equity; surveys by Caton \textit{et al.}~\cite{CatonH24} confirm that both strategies leave measurable residual bias.  With the rise of large language models, Hua \textit{et al.}~\cite{HuaGXJLZ24} attach a counterfactual discriminator to a pre-trained encoder, Deldjoo \textit{et al.}~\cite{cfairllm2024} propose benchmark protocols for consumer fairness, and Zhang \textit{et al.}~\cite{ZhangBZWF023} diagnose systematic bias in zero-shot ChatGPT recommendations; complementary surveys by Gallegos \textit{et al.}~\cite{Gallegos2024BiasSurvey} trace broader LLM biases, while Rarrick \textit{et al.}~\cite{Rarrick2024GateXE} release challenge sets that surface gender disparity in translation.  Our study extends this line by introducing a kernel-lifted projection and gated adapter that eliminate multi-attribute leakage in fine-tuned LLM recommenders without additional trainable parameters, thereby complementing earlier re-ranking and adversarial methods. 

\textbf{Sub-space Projection for Debiasing}\label{sec:subspace}
Projection methods first \emph{locate} the directions that encode a protected attribute and then \emph{drop} those directions by orthogonal projection.  The idea began with static word embeddings: Bolukbasi~\textit{et~al.}~\cite{BolukbasiCZSK16}\ showed that subtracting a single ``gender direction''—estimated from pairs such as \textit{he–she}—reduces analogy bias in word2vec vectors~\cite{abs-1301-3781}.  Iterative Null-space Projection (INLP) extends this trick to contextual encoders by repeatedly (i) training a linear classifer and (ii) projecting representations onto its null-space until the attribute is no longer linearly predictable~\cite{RavfogelEGTG20}.  Because linear classifer can miss non-linear leakage, later work performs the projection in a reproducing-kernel Hilbert space and approximates it with random Fourier features (RFF), so that \emph{kernel-linear} signals are removed as well~\cite{RavfogelTGC22,RahimiR07}.  Most recently, Cho~\textit{et~al.}\ connect this ``project-then-repair'' view to information-theoretic sufficiency, showing that a representation can stay predictive for the downstream label yet be conditionally independent of the sensitive attribute~\cite{0001MGM24}.

\section{Preliminaries}
\label{sec:prelims}


\subsection{LLM-based Recommender Systems}
Let $\mathcal{U}$ be the set of users, $\mathcal{I}$ the set of items, and $\mathcal{C}$ the set of contextual information, such as user interaction history, item metadata, or natural language queries. Given a large language model $\mathcal{M}$ parameterized by $\theta$, the objective of an LLM-based recommender system (RecLLM) is to learn a mapping: $
f_{\theta} : (\mathcal{U}, \mathcal{I}, \mathcal{C}) \rightarrow \mathbb{R}$ 
that assigns a relevance score or generates a ranked list of items $\{i_1, i_2, \ldots, i_k\} \subseteq \mathcal{I}$ for a user $u \in \mathcal{U}$, conditioned on context $c \in \mathcal{C}$. The function $f_\theta$ operates via prompt generation. The objective is to optimize $f_\theta$ such that recommendations align with user intent and context while leveraging the generalization capabilities of LLMs.


A typical RecLLM accepts a prompt as an input, and replies with one item identifier. 
In this paper, we study two different types of  prompts for RecLLMs~\cite{HuaGXJLZ24,Geng0FGZ22}. Figure~\ref{fig:prompt-exmp} illustrates an example of both types of recommendations.

\textbf{Direct recommendation.} The prompt names the user and lists a set of candidate items, without showing any user-item historical interactions. The LLM inspects the prompt and outputs a single item for recommendation.  
\begin{wrapfigure}{r}{0.48\textwidth}  
  \vspace{-10pt}                        
  \begin{tcolorbox}[
      colback=gray!3,
      boxrule=0.2pt,
      left=2pt,right=2pt,top=2pt,bottom=2pt,
      width=0.47\textwidth               
  ]
  \small\setlength{\parskip}{0pt}
  \textbf{Direct Recommendation}\\
  \bera{\textcolor{blue}{Input}: Which movie user\_\{user\_ID\} would like to watch among the following candidates? \{List of 100 IDs\}.}\\
  \bera{\textcolor{orange}{Output}: \{movie\_ID\}}\\
  \vspace{3pt}
  \textbf{Sequential Recommendation}\\
  \bera{\textcolor{blue}{Input}: User\_\{user\_ID\} has already watched the following movies \{sequence of movie IDs\}.}\\
  \bera{Which movie user\_\{user\_ID\} would like to watch next?}\\
  \bera{\textcolor{orange}{Output}: \{movie\_ID\}}
  \end{tcolorbox}
  \caption{Prompt templates for LLM-based recommendations.}
  \label{fig:prompt-exmp}
  \vspace{-2.5em} 
\end{wrapfigure}
\textbf{Sequential recommendation.} The prompt includes a sequence of user’s recent  interacted items associated with specific users. The LLM outputs the recommended item based on the historical interactions. 

In RecLLMs, the goal is to convert structured recommendation data (e.g., user-item interactions, ratings, timestamps) into text sequences that the model can interpret and reason over.

Given its complexity, it is difficult to train a RecLLM from scratch. Therefore, in this paper, we only consider fine-tuning of RecLLMs with \emph{Low-Rank Adaptation} (LoRA)~\cite{HuSWALWWC22}, which introduces a low-rank decomposition to represent the parameter updates by fine-tuning. 
We call each such update a ``LoRA patch''. Throughout the paper, we only consider training of LoRA patches while the backbone of the RecLLM model stays frozen.

\subsection{Counterfactual Fairness}
In this paper, we follow the definition of {\em counterfactual fairness}~\cite{KusnerLRS17}:
A recommender system is fair for an individual if its predicted recommendations would remain
unchanged were the person assigned any other demographic label.  
Unlike group-level fairness metrics such as demographic parity~\cite{dp-data-08}, 
which can yield unstable estimates in the presence of severe class imbalance~\cite{transbalance24}, 
the counterfactual fairness is defined at the individual level and thus is robust to the skewed distribution. As a result, it has been increasingly adopted to the LLM setting as a standard for fine-grained fairness assessment.

\begin{definition}[Counterfactually Fair Recommendations \cite{KusnerLRS17}]\\
Let $A\!\in\!\{a_1,\dots,a_m\}$ be a sensitive attribute and let 
$\mathbf h_u$ denote the part of the user representation that is  
causally independent of $A$.  
Write $\mathbf L(\mathbf h_u,A)$ for the (possibly stochastic) top-$K$  
list returned by the recommender.  
The recommender is \emph{counterfactually fair} iff
\begin{align*}
    P\!\bigl(\mathbf L(\mathbf h_u,A)=\ell \mid A=a_i\bigr)
    =
    P\!\bigl(\mathbf L(\mathbf h_u,A)=\ell \mid A=a_j\bigr),
    \forall\ell,\;\forall a_i,a_j\in\{a_1,\dots,a_m\}.&
\end{align*}

In other words, an individual would receive exactly the same distribution of
recommendations had their demographic label been any other feasible
value.  
\end{definition}

In practice,
enumerating \emph{all} counterfactual prompts  is infeasible. Inspired by Hua \textit{et al.}~\cite{HuaGXJLZ24}, we
adapt the protocol such that: we freeze the
LLM’s sequence-level representation $\mathbf{h}\in\mathbb{R}^{d}$  and train a two-layer MLP 
$g_{\theta}\!:\mathbb R^{d}\!\to\!\mathbb R^{m}$ to predict a sensitive attribute $A\in\{a_{1},\dots ,a_{m}\}$. 
For each one-vs-rest setting, we compute the MLP’s AUC, measured as the probability that its score for a randomly chosen user in class $a_{i}$ exceeds the score for an independently drawn user outside the class: 
\begin{equation}
    \operatorname{AUC}_i = \operatorname{AUC}\ \!\bigl(g_{\theta}(\mathbf h),\,A=a_{i}\text{ vs. }A\neq a_{i}\bigr),
\quad
i=1,\dots,m.
\end{equation}

The bias of the sensitive attribute $A$  can be measured by the \textbf{Counterfactual Leakage Gap (CLG)} on $A_i$, defined as the average deviation of the AUC scores across all classes of 
$A$ from the random-guessing baseline (i.e., AUC = 0.5). 
\begin{equation}
    \label{eq:clg}
\Delta_{\operatorname{CL}}\!=\!\frac1m\sum_{i=1}^{m}|\operatorname{AUC}_i-0.5| 
\end{equation}

Intuitively, the perfect counterfactual fairness on the sensitive attribute $A$ is reflected by $\Delta_{\operatorname{CL}}=0$. A high  $\Delta_{\operatorname{CL}}$ value reflects greater bias in the recommendations concerning the attribute $A$.

If the model representations are independent from the sensitive information, then the recommendations should be bias free (i.e., $\Delta_{\operatorname{CL}}=0.5$). 
Therefore, the main objective of this paper is to learn fair data
representations that are i) informative enough for the downstream task, and ii)
independent from the sensitive attributes. 
Formally, let $\mathbf h$ be the sequence-level representation learned by the given RecLLM model. Our goal is to convert $\mathbf h$ to $\mathbf h^\star$ such that $\mathbf h^\star$ is independent of $A$.

\section{Method}
\label{sec:method}

Given the sequence-level representation
$\mathbf h\!\in\!\mathbb R^{d}$, a $d$-dimensional vector that
summarizes each prompt, we perform the following two steps to remove the bias encoded in $\mathbf h$.
By Step 1, we erase the information of a single sensitive attribute encoded in $\mathbf h$. 
Step 2 extends Step 1 to removing bias from multiple sensitive attributes while minimizing the accuracy loss of the recommendations.

\textbf{Step 1: Debiasing on a single sensitive attribute.}
 This step takes the sequence-level representation $\mathbf h$ of the RecLLM model and a specific sensitive attribute $A$ as the input, and transforms $\mathbf h$ to $\mathbf h^\star$ which is independent from $A$.  To capture the non-linear signal in the representations, $\mathbf h$ is first lifted to a higher dimension. Then the lifted $\mathbf h$ is projected to $\mathbf h^\star$ by INLP. 
 
\textbf{Step 2: Dealing with multiple sensitive attributes.}
When multiple sensitive attributes are present, independently generating a fair INLP representation for each attribute using Step 1 can lead to significant degradation in recommendation accuracy. To address this challenge, we propose a {\em two-level gated LoRA adapter} that introduces two gating mechanisms. The first gate produces a soft weighting vector that adaptively scales the INLP projector associated with each sensitive attribute, enabling more nuanced debiasing. The second gate modulates a LoRA-based patch applied to the debiased representation, allowing for lightweight fine-tuning that compensates for the accuracy loss introduced by the projection step.


\subsection{Debiasing on a Single Sensitive Attribute}

\label{sec:rff-inlp}

Intuitively, applying INLP directly on the sequence-level representation can remove the bias of a given sensitive attribute. However, such projection is only able to produce linear representations and may lose the non-linear signals exist in LLM models.  
To address this issue, we conduct the following steps: 
(1) {\bf Random Fourier Lift}: The representation vector is lifted into a random-Fourier feature space~\cite{RahimiR07} whose inner product unbiasedly approximates the Gaussian kernel to make the bias become more distinct for removal; 
(2) {\bf INLP projection}: The representation in the lifted space is debiased by INLP; 
(3) {\bf Refinement of projection}: the projection is further refined, aiming to recover the non-linear signals for LLMs to utilize. 
Fig.~\ref{fig:single-pipeline} illustrates the flow of the three steps. 
Next, we explain the details of each step.


\smallskip
\begin{wrapfigure}{r}{0.45\textwidth}
 \vspace{-2.5em}
  \centering
  \includegraphics[
    width=\linewidth,
    trim={20pt 30pt 40pt 30pt}, 
    clip
  ]{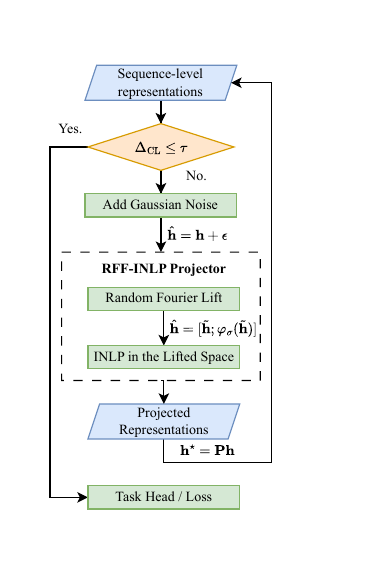}
  \caption{Debiasing on a sensitive attribute using RFF-INLP.}
  \label{fig:single-pipeline}
  \vspace{-1.5em}
\end{wrapfigure}
\noindent\textbf{Step 1: Random Fourier Lift.}
The goal of this step is to re-express representation in a way that makes any subtle bias to be more distinct for an easier removal later. 
Given the sequence-level representation $\mathbf h\!\in\!\mathbb R^{d}$, we first add an 
isotropic perturbation $\epsilon$ to $\mathbf h$, aiming to prevent INLP from ignoring low-variance directions: 
\[
\epsilon\sim\mathcal N(\mathbf 0,\eta^{2}\mathbf I_{d}),\qquad
\tilde{\mathbf h}=\mathbf h+\epsilon ,
\]
Here, $\epsilon$ is a $d$-dimensional Gaussian noise vector with  mean $\mathbf{0}$ and covariance $\eta^{2}\mathbf{I}$, where $\eta>0$ controls the noise magnitude and $\mathbf I$ is the $d\times d$ identity matrix. The isotropic perturbation adds the same tiny variance $\eta^{2}$ to every coordinate, raising low-variance dimensions to the same scale as the dominant ones. Once this variance is equalized, any remaining attribute signal can no longer hide in small-amplitude directions \cite{eftekhari2025importance}. 

Next, we draw
\[
   \mathbf\Omega \sim \mathcal N(\mathbf 0,\sigma^{-2}\mathbf I_{d}),
   \qquad
   \mathbf b \sim \mathrm{Unif}[0,2\pi]^{D},
\]
where $\mathbf\Omega\in\mathbb R^{D\times d}$ stores $D$ ``frequency'' vectors drawn from the Gaussian law; $\mathbf b\in\mathbb R^{D}$ supplies independent phase shifts sampled uniformly from $[0,2\pi]$.
We then map the balanced representation $\tilde{\mathbf h}$ into an $D$-dimensional feature vector using the function $\varphi_{\sigma}$: 
\begin{align*}
    \varphi_{\sigma}(\tilde{\mathbf h})
    &=\sqrt{\tfrac{2}{D}}\,
     \cos(\mathbf\Omega\tilde{\mathbf h}+\mathbf b)\in\mathbb R^{D}
\end{align*}
whose inner product unbiasedly approximates the Gaussian kernel $k_{\sigma}$. 
Finally, the random-Fourier feature is generated as follows~\cite{RahimiR07}
\begin{align*}
    \hat{\mathbf h}&=[\,\tilde{\mathbf h};\varphi_{\sigma}(\tilde{\mathbf h})\,]\in\mathbb R^{d+D},
\end{align*}
where $[\cdot; \cdot]$ is a concatenation operation. We call $\hat{\mathbf{h}}$ the RFF-augmented features.

A nice property of RFF-augmented features is that, taking the dot product of two feature vectors (e.g., $\tilde{\mathbf h},\tilde{\mathbf h}'$) already gives (in expectation) the same value as their Gaussian-kernel similarity \cite{RahimiR07,ObliviousData2021}, which means
\begin{equation}
    \label{eq:rff-expectation}
    \mathbb E_{(\mathbf\Omega,\mathbf b)}\! \bigl[  \langle\varphi_{\sigma}(\tilde{\mathbf h}),
              \varphi_{\sigma}(\tilde{\mathbf h}')\rangle]   =k_{\sigma}(\tilde{\mathbf h},\tilde{\mathbf h}'),
              \quad\forall\,\tilde{\mathbf h},\tilde{\mathbf h}'\in\mathbb R^{d}.
\end{equation}  
The INLP probes trained on these coordinates therefore perceive - and subsequently remove - every kernel-linear leakage direction that was invisible in the original feature space.

\smallskip
\noindent\textbf{Step 2: INLP Projection.}
\label{subsec:inlp_cal}
Let $\{\hat{\mathbf h}^{(n)}\}_{n=1}^{N}\in\mathbb R^{d+D}$ be the RFF 
representations output by Step 1, where $N$ is the size of the samples.  
INLP iteratively trains a sequence of linear probes 
$\hat{w}_{1},\dots,\hat{w}_{T}\in\mathbb R^{d+D}$ by $T$ iterations
that each maximizes the conditional log-likelihood of the sensitive
attribute on the current sub-space, then removes that direction.  
Stacking row-wise the probe weights gives the matrix

\[
\hat{\mathbf W}
  =\begin{bmatrix}
      \hat{w}_{1}^{\!\top}\\[-2pt]
      \vdots\\[-2pt]
      \hat{w}_{T}^{\!\top}
    \end{bmatrix}
  \in\mathbb R^{T\times(d+D)},
\]
where T is the number of iterations for computing the linear probes and we assume that $T\leq (d+D)$. 
Each probe will first compute the projector $P$, then we test the projected representation whether it passes the counterfactual fairness threshold. If not, we continue to compute another linear probe. After $T$ times we can obtain a representation satisfying the fairness threshold.
The final projector is obtained by orthogonally removing the row space of $\hat{\mathbf W}$. For a full-row-rank matrix this projector has the classical closed form~\cite{golub2013matrix}

\[
\hat{\mathbf P}
  =\mathbf I_{d+D}
   -\hat{\mathbf W}^{\!\top}
    (\hat{\mathbf W}\hat{\mathbf W}^{\!\top})^{-1}
    \hat{\mathbf W},
\quad
\hat{\mathbf P}^{2}=\hat{\mathbf P}^{\!\top}=\hat{\mathbf P},
\]

where $\hat{\mathbf P}$ is the orthogonal projection onto the
null-space of $\hat{\mathbf W}$ \cite{RavfogelEGTG20}. 
Downstream layers still expect a $d$-dimensional vector, so we must
extract the part of the full projector $\hat{\mathbf P}$ that acts on
those $d$ coordinates.  We therefore write $\hat{\mathbf P}$ in block
form

\[
\hat{\mathbf P}
  =\begin{bmatrix}
      \hat{\mathbf P}_{11} & \hat{\mathbf P}_{12}\\[2pt]
      \hat{\mathbf P}_{21} & \hat{\mathbf P}_{22}
    \end{bmatrix},
\qquad
\hat{\mathbf P}_{11}\in\mathbb R^{d\times d},
\]

We only keep the upper-left block $\hat{\mathbf P}_{11}$ as the projection result, following the well-known \emph{Schur-complement} identity~\cite{golub2013matrix} for
orthogonal projectors: it tells us that the action of
$\hat{\mathbf P}$ on the first $d$ coordinates is captured entirely by $\hat{\mathbf P}_{11}$. 
The move is purely for dimension matching and does not itself influence bias removal; it simply lets the debiased vector flow into the original network without changing its size. Following that, we have
\[
\mathbf{h^\star}=\mathbf P\,\mathbf{h}
  =
  \hat{\mathbf P}_{11}\mathbf{h},
\qquad
\mathbf P=\hat{\mathbf P}_{11}\in\mathbb R^{d\times d}.
\]

Here, $\mathbf{h^\star}$ is the debiased representation. Notably, as the matrix $\mathbf P$ inherits symmetry and idempotence from
$\hat{\mathbf P}$, 
all directions spanned by the original probes
$\{\hat{w}_{t}\}_{t=1}^{T}$ intersecting the $\mathbf h$-subspace
have been removed.

\smallskip
\noindent\textbf{Step 3: Refinement of Projection.}
At the end of every training epoch of the LLM model, we compute the Counterfactual Leakage Gap
$\Delta_{\mathrm{CL}}$ (Eqn.~\eqref{eq:clg}).
We use a parameter, $\tau\in(0, 1)$, to control the degree of fairness. A smaller $\tau$ 
value requires stricter fairness. 
Then, we repeat Steps 1 \& 2 to update
$\mathbf P$ and $\mathbf{h}$ if $\Delta_{\mathrm{CL}}\!>\!\tau$, 
and retain the current $\mathbf P$ for the next training epoch otherwise. 
Our empirical evaluations shows that no more than three updates needed to achieve a good performance, indicating the efficiency of our approach. 


\subsection{Handling Multiple Sensitive Attributes}
\label{sec:gated-moe}

Debiasing multiple sensitive attributes with the INLP projectors 
raises two problems. First, each attribute shows up in the representation with a different intensity. For example, the {\em gender} attribute is usually  predicted from only a few components of the representations, whereas the {\em age} and {\em  occupation} attributes  spread across many more components. In this case, using one fixed projection for all of them risks removing too much useful information for gender or, conversely, removing too little bias for age and occupation. 
Second, these attributes are often correlated in the embedding: the same components that help the model guess a user’s age can also capture general preferences shared across all users. If we zero-out those correlated dimensions to reduce age bias, this will incur high loss of the recommendation accuracy.
We address both issues by designing a lightweight {\em two-level gated MoE adapter}, which is inserted after every self-attention block of the backbone of LLMs. 
The adapter comprises two gates: (1) A {\em context-dependent outer gate} softly sets, for each attribute, how much of its projector to mix in, sliding from zero to full strength; 
and (2) An {\em inner gate} adds a low-rank LoRA patch, based on the already-debiased vector, to bring back only the features that improve accuracy without re-introducing sensitive information. Together the gates form an ``erase-then-repair'' cycle that handles all attributes at once without duplicating the model or adding extra optimizers.

For layer $\ell$ and token position $t$, let
$\mathbf h_{\ell,t}\!\in\!\mathbb R^{d}$ be the token-level hidden state immediately before the layer’s pre-trained output projection $\mathbf O_{\ell}\!\in\!\mathbb R^{d\times d}$.
A pooled user-context vector
$\mathbf c_u=\operatorname{Pool}\!\left(\{\mathbf h_{\ell,t}\}_{t}\right)\in\mathbb R^{d}$
is the input for the gating functions described below; 
intuitively, $\mathbf c_u$ is a single $d$ dimensional summary of the entire prompt, obtained by averaging (or attention-weighting) all token hidden states at layer~$\ell$. It captures the overall user context in one compact vector.
Each adapter instantiation maintains $K$ attribute-specific experts, implemented as rank-$r$ LoRA updates.

\paragraph{Level-1 soft projection (attribute weighting)}

Given the pooled user context $\mathbf c_u\!\in\!\mathbb R^{d}$ we first
compute a \emph{contextual attention vector}
\[
    \boldsymbol\alpha
        = \operatorname{softmax}\!\bigl(\mathbf G_1,\mathbf c_u\bigr)
        \quad\in\Delta^{K},
    \qquad
    \mathbf G_1\in\mathbb R^{K\times d}.
\]
Here $\Delta^{K}\!=\!\{\mathbf p\in\mathbb R^{K}\mid p_k\!\ge\!0,\,
\sum_{k=1}^{K}p_k\!=\!1\}$ designates the $K$-dimensional
\emph{probability simplex}; its dummy variable $\mathbf p$ is replaced
in practice by $\boldsymbol\alpha=(\alpha_1,\dots,\alpha_K)$.  
The matrix $\mathbf G_1$ is the \emph{only} trainable component at this
level; its $k$-th row acts as a key that measures how salient the
$k$-th sensitive attribute is in the current user trace.  
Each attribute owns a hard projector
$\mathbf P^{(k)}\!\in\!\mathbb R^{d\times d}$, computed once in Subsection~\ref{subsec:inlp_cal}, that is orthogonal and idempotent
($\mathbf P^{(k)}{=}\mathbf P^{(k)}\mathbf P^{(k)}$) and removes the
linear signal of attribute $A_k$.
Because $\boldsymbol\alpha$ lies on the simplex, it serves two roles:
\begin{enumerate}
  \item \textbf{Adaptive attribute weighting.}
        A larger coefficient $\alpha_k$ indicates stronger residual leakage for attribute $A_k$; hence a larger correction is applied, whereas already ``clean'' attributes receive near-zero weights and incur negligible utility loss.
  \item \textbf{Controlled projection.}
        The softly projected representation
        $\mathbf{h}^{\star}
          =\bigl(\mathbf I_d-\sum_{k}\alpha_k(\mathbf I_d-\mathbf P^{(k)})\bigr)\mathbf h$
        depends on a convex combination, whose Jacobian
        $\tfrac{\partial\mathbf{h}^{\star}}{\partial\mathbf h}
          =\mathbf I_d-\sum_{k}\alpha_k(\mathbf I_d-\mathbf P^{(k)})$
        remains full rank whenever at least one $\alpha_k<1$,
        thereby avoiding catastrophic loss of expressivity.
\end{enumerate}

The softly projected representation fed to the second gate is a convex combination of the original representations $\mathbf h$ and its attribute-wise hard-projected versions $\mathbf P^{(k)}\mathbf h$:
\[
\mathbf{h}^{\star}
   =\mathbf h+\sum_{k=1}^{K}\alpha_k\!\bigl(\mathbf P^{(k)}\mathbf h-\mathbf h\bigr)
   =\Bigl(\mathbf I_d-\sum_{k=1}^{K}\alpha_k(\mathbf I_d-\mathbf P^{(k)})\Bigr)\mathbf h .
\]
This construction exposes the backbone to a \emph{continuum} that spans the untouched feature ($\alpha_k=0$) and the fully projected one ($\alpha_k=1$), allowing the network to find an information–fairness trade-off in between.  The update remains \emph{residual}: because $\mathbb E[\alpha_k]=\tfrac1K$ under a uniform prior, we have $\mathbb E[\mathbf h^\star\mid\boldsymbol\alpha]=\mathbf h$, so gradients flow through the adapter as stably as through a standard skip-connection—an empirical necessity when stacking many transformer layers.

\paragraph{Level-2 attribute-conditioned expert fusion}

Soft projection suppresses most sensitive cues yet inevitably removes
some task‐useful variation.  To restore that capacity we attach, to each
sensitive attribute $A_k (k\!\in\!\{1,\dots,K\})$, a rank-$r$ LoRA expert
that perturbs the \emph{pre-trained output projection}
$\mathbf O_{\ell}$ of the current layer in the RecLLM:
\[
   \Delta\mathbf O_{k}
        = \mathbf U_{k}\,\mathbf V_{k},
   \quad
   \underbrace{\mathbf V_{k}\in\mathbb R^{r\times d}}_{\text{rank-$r$ down-projection}}\!
   \
   \underbrace{\mathbf U_{k}\in\mathbb R^{d\times r}}_{\text{up-projection to hidden size}}
\]

so that $\mathbf O_{\ell}+\Delta\mathbf O_{k}$ spans the subspace most
relevant to attribute $A_k$. The contribution of this expert is gated
by a \emph{sigmoid weight}
\[
   \beta_{k}
        = \sigma\!\bigl(\mathbf g_{k}^{\!\top}\,\tilde{\mathbf c}_u\bigr),
   \qquad
   \mathbf g_{k}\in\mathbb R^{d},
\]
where $\tilde{\mathbf c}_u$ is a LayerNorm-stabilised summary of the
soft-purified sequence.  Because $\beta_{k}\in(0,1)$, the gate acts as
an analogue switch: it closes when leakage for attribute $A_k$ has been
eliminated and opens proportionally when a modest utility-bearing signal
remains.

The token representation finally forwarded to the next transformer block
is
\[
   \mathbf y
     = \mathbf O_{\ell}\mathbf{h}^{\star}
       + \sum_{k=1}^{K}
           \beta_{k}\,
           \Delta\mathbf O_{k}\,
           \mathbf{h}^{\star},
\]
where the first term preserves debiased backbone semantics and the
second term selectively re-injects attribute-specific nuance.  Each
$\Delta\mathbf O_{k}$ being rank $r$ limits the additional footprint
to $O(Kdr)$ parameters and flops; sharing $\mathbf U_{k}$ across
layers further reduces memory.

To ensure the gates make clear yet well-balanced choices, we add two regularizers:
(i) an \emph{entropy penalty} on $\boldsymbol\alpha$ to prevent the
level-1 gate from collapsing to a premature one-hot vector, and
(ii) an $\ell_{1}$ penalty on the vector
$\boldsymbol\beta=(\beta_{1},\dots,\beta_{K})$ to promote sparse
expert activation.  Together, the two gating layers realise an
attribute-aware \emph{decouple–repair} mechanism: the first gate
attenuates unfair directions while preserving rank, and the second gate
injects low-rank corrections only when they do not re-introduce
sensitive-attribute information.

\subsection{Training and Inference}

We train two types of LoRA adapters simultaneously: 
(1) {\bf Task-LoRA adapters} are inserted into the linear layers that project queries, keys, values, and outputs inside each self-attention block of the LLM's backbone,  and 
(2) {\bf Attribute-LoRA
experts} reside inside the level-2 gate and
are scaled by \(\beta_{k}=\sigma(\mathbf g_{k}^{\top}\tilde{\mathbf
c}_{u})\). These experts are applied only to the output-projection layer. Both types of adapters receive gradients solely from the token-level
negative log-likelihood 
\[
\mathcal L_{\text{task}}
  = -\log P_{\theta}\!\bigl(i^{\star}\mid x_{u}\bigr),
\]
where \(P_{\theta}\) is the backbone’s predictive distribution, \(x_{u}\) is the full text prompt that encodes the target user’s historical interactions, and $i^{\star}$ is the ground-truth item. No additional entropy or sparsity terms is needed.

Gradient updates only happen in the adapter parameters
\(\{\mathbf G_{1},\mathbf g_{k},\mathbf U_{k},\mathbf V_{k}\}_{k=1}^{K}\)
and the task-LoRA weights. 
The orthogonal projectors
\(\mathbf P^{(k)}\) and their composite \(\mathbf P^{\star}\) remain
frozen, and are updated  only when the leakage probe detects
\(\Delta_{\mathrm{CL}}>\tau\) (Section~\ref{subsec:inlp_cal}).

\section{Experimental Evaluation}

We present empirical results on two real-world benchmarks encompassing both direct and sequential recommendation. Our INLP Projection + Two-Level MoE Adapter, built on an Instruct Llama-3 backbone enhanced with LLaRA-style~\cite{liao2024llara} low-rank adapters, is evaluated against the fairness-oriented baseline UP5~\cite{HuaGXJLZ24}. We report utility (Hit@{1,3,10}) alongside fairness, quantified by the Counterfactual Leakage Gap $\Delta_{\mathrm{CL}}$.

\subsection{Experimental Setup}\label{sec:setup}

\begin{wraptable}{r}{0.5\linewidth}  
  \vspace{-6pt}
  \centering
  \caption{Performance of the backbone approach on \textbf{MovieLens} and \textbf{Insurance} datasets.
Utility: Hit rate (\%$\uparrow$); fairness: counterfactual–leakage gap $\Delta_{\mathrm{CL}}$ (\%$\downarrow$).}
  \label{table:backbone}
    \small
    \setlength{\tabcolsep}{3.5pt}
    \renewcommand{\arraystretch}{1.05}
    \begin{adjustbox}{max width=\linewidth}
    \begin{tabular}{lSSS}
    \toprule
    & \multicolumn{2}{c}{\textbf{MovieLens}} & {\textbf{Insurance}}\\
    \cmidrule(lr){2-3}\cmidrule(lr){4-4}
    LLaRA& {Sequential} & {Direct} & {Direct}\\
    \midrule
    Hit@1 $\uparrow$  & 58.03 & 45.20 & 57.66\\
    Hit@3 $\uparrow$  & 76.58 & 60.20 & 91.24\\
    Hit@10 $\uparrow$ & 82.66 & 71.20 & 99.10\\
    \cmidrule{1-4}
    $\Delta_{\mathrm{CL}}$ Gender $\downarrow$   & 21.90 & 29.80 &  0.90\\
    $\Delta_{\mathrm{CL}}$ Marital $\downarrow$  & {--}  & {--}  & 11.20\\
    $\Delta_{\mathrm{CL}}$ Age $\downarrow$      & 19.10 & 18.40 &  1.90\\
    $\Delta_{\mathrm{CL}}$ Occup. $\downarrow$   & 12.40 & 11.00 &  3.40\\
    $\Delta_{\mathrm{CL}}$ Avg. $\downarrow$     & 17.80 & 16.40 &  4.35\\
    \bottomrule
  \end{tabular}
  \end{adjustbox}
  \vspace{-6pt}
\end{wraptable}
\textbf{Datasets.}  
We use two datasets in our experiments: 
(1) MovieLens-1M data~\cite{HarperK16}, which is a timestamped movie-rating data collected by the GroupLens research group; 
(2) Insurance dataset\footnote{\hyperlink{https://www.kaggle.com/datasets/mrmorj/insurance-recommendation}{https://www.kaggle.com/datasets/mrmorj/insurance-recommendation}} drawn from an African insurer’s customer portfolios. 
We consider the following sensitive attributes for the MovieLens dataset:  {\em Gender} (2 classes), {\em Age} (7 classes), and {\em Occupation}  (21 classes). 
After sorting each user’s ratings chronologically, we train on all but the last two interactions, validate on the penultimate one, and test on the final one.
We consider the following sensitive attributes for the Insurance dataset: {\em Marital status} (8 classes), {\em Age} (5 classes), and {\em Occupation} (6 classes). 
To train the LLM, we sort the interactions in the Insurance dataset by date, using the prefix up to the third-last item interaction for training, the second-last item for validation, and the last one for testing.

\textbf{Evaluation metrics.}  
Utility is assessed with Hit@$\{1,3,10\}$, computed on the held-out target item given a candidate set of one positive and 99 sampled negatives; Hit@$k$ records whether the true item appears in the top-$k$ ranks.  Fairness is measured by the \emph{Counterfactual Leakage Gap} (CLG) $\Delta_{\operatorname{CL}}$ in Eq.~\ref{eq:clg}: after fine-tuning we freeze the sequence-level representation $\mathbf h$, train a two-layer MLP probe for each sensitive attribute, compute one-vs-rest AUCs, subtract the random-guess baseline 0.5 and average the deviations over all classes; the closer this average is to zero, the less linearly recoverable sensitive information remains. We compute the average performance when multiple sensitive attributes are considered. 

\textbf{LLM backbone.}  
All of our experiments are conducted within the RecLLM framework based on LLaRA~\cite{liao2024llara}, using an Instruct Llama-3.2\footnote{\hyperlink{https://huggingface.co/datasets/meta-llama/Llama-3.2-1B-Instruct-evals}{https://huggingface.co/datasets/meta-llama/Llama-3.2-1B-Instruct-evals}} 1B model as the frozen base encoder–decoder.  On this backbone we add two debiasing components only: (i) a fixed RFF–INLP projector with $D=4096$ random Fourier features and a noise-refine scale $\eta=0.05$; and (ii) a two-level gated MoE adapter whose task-side LoRA updates have rank 32 and whose attribute-specific experts have rank 8. These hyper-parameters follow the public LLaRA configuration~\cite{liao2024llara} and are kept identical for every dataset and baseline.

\textbf{Baselines.}
For comparison we include (i) the standard LLaRA~\cite{liao2024llara} model without any debiasing modules, (ii) P5~\cite{Geng0FGZ22}, which adapts the backbone through prompt engineering, and (iii) UP5~\cite{HuaGXJLZ24}, a fairness-oriented extension of P5.  Reported scores for P5 and UP5 follow the configurations in their original study, ensuring that all methods are evaluated under the same candidate-sampling protocol and metric definitions.

\subsection{Performance Evaluation}

\begin{table}[t]
\centering
\caption{\textbf{Sequential recommendation} results on {\bf MovieLens} dataset. Utility metric: Hitting rate (\%$\uparrow$). Fairness metric: counterfactual-leakage gap $\Delta_{\operatorname{CL}}$ (\%$\downarrow$). Bold = best.}
\label{table:movie_sqe}
\setlength{\tabcolsep}{4pt}
\renewcommand{\arraystretch}{1.05}

\begin{adjustbox}{max width=\textwidth}
\begin{tabular}{lcccccccc}
\toprule
\multirow{2}{*}{Sensitive} 
  & \multicolumn{2}{c}{Hit@1 $\uparrow$} 
  & \multicolumn{2}{c}{Hit@3 $\uparrow$} 
  & \multicolumn{2}{c}{Hit@10 $\uparrow$} 
  & \multicolumn{2}{c}{$\Delta_{\mathrm{CL}}$ $\downarrow$} \\
\cmidrule(lr){2-3}\cmidrule(lr){4-5}\cmidrule(lr){6-7}\cmidrule(lr){8-9}
attribute& UP5 & Ours & UP5 & Ours & UP5 & Ours & UP5 & Ours \\
\midrule
Gender (G)     & 26.82 & \textbf{39.69} & 45.18 & \textbf{54.45} & 64.38 & \textbf{64.93} & 4.19 & \textbf{0.90} \\
Age (A)      & 31.23 & \textbf{38.16} & 51.18 & \textbf{53.61} & \textbf{67.70} & 65.13 & 2.91 & \textbf{0.60} \\
Occupation (O)  & 31.66 & \textbf{38.58} & 50.73 & \textbf{54.02} & \textbf{67.45} & 65.22 & \textbf{0.00} & 1.70 \\
G + A + O & --    & \textbf{56.08} & --    & \textbf{72.28} & --    & \textbf{81.67} & --   & \textbf{0.17} \\
\bottomrule
\end{tabular}
\end{adjustbox}
\end{table}


\begin{table}[t]
\centering
\caption{\textbf{Direct recommendation} results on \textbf{MovieLens} and \textbf{Insurance} datasets. 
Utility metric: Hit rate (\%$\uparrow$). Fairness metric: counterfactual-leakage gap $\Delta_{\operatorname{CL}}$ (\%$\downarrow$).  
Bold = best.}
\label{table:direct_combined}
\setlength{\tabcolsep}{4pt}
\renewcommand{\arraystretch}{1.05}

\begin{adjustbox}{max width=\textwidth}
\begin{tabular}{llcccccccc}
\toprule
\multirow{2}{*}{\textbf{Dataset}} & \multirow{2}{*}{\textbf{Attribute}} 
& \multicolumn{2}{c}{Hit@1 $\uparrow$} 
& \multicolumn{2}{c}{Hit@3 $\uparrow$} 
& \multicolumn{2}{c}{Hit@10 $\uparrow$} 
& \multicolumn{2}{c}{$\Delta_{\mathrm{CL}}$ $\downarrow$} \\ 
\cmidrule(lr){3-4}\cmidrule(lr){5-6}\cmidrule(lr){7-8}\cmidrule(lr){9-10}
& & UP5 & Ours & UP5 & Ours & UP5 & Ours & UP5 & Ours \\
\midrule
\multirow{4}{*}{MovieLens}
  & Gender (G)     & 16.38 & \textbf{36.92} & 35.04 & \textbf{48.28} & 65.82 & \textbf{73.01} & 4.19 & \textbf{0.60} \\
  & Age (A)        & 21.22 & \textbf{37.14} & 39.22 & \textbf{49.35} & 67.30 & \textbf{69.72} & 2.91 & \textbf{0.00} \\
  & Occupation (O)  & 21.00 & \textbf{34.60} & 38.50 & \textbf{46.57} & \textbf{69.00} & 64.34 & \textbf{0.00} & 0.60 \\
  & G + A + O & 20.18 & \textbf{43.20} & 38.79 & \textbf{49.42} & \textbf{66.78} & 66.06 & 3.21 & \textbf{0.78} \\
\midrule
\multirow{4}{*}{Insurance}
  & Marital (M)     & \textbf{81.03} & 54.23 & 90.58 & \textbf{93.39} & 97.66 & \textbf{99.16} & 2.19 & \textbf{1.30} \\
  & Age (A)         & \textbf{82.53} & 55.59 & 92.68 & \textbf{93.45} & 98.89 & \textbf{99.04} & \textbf{0.00} & 0.50 \\
  & Occupation (O) & \textbf{82.53} & 56.08 & 82.68 & \textbf{94.04} & 98.89 & \textbf{99.26} & 3.28 & \textbf{1.90} \\
  & M + A + O & \textbf{81.63} & 57.37 & \textbf{91.52} & 91.47 & 97.37 & \textbf{99.35} & 0.74 & \textbf{0.13} \\
\bottomrule
\end{tabular}
\end{adjustbox}
\end{table}

Table \ref{table:backbone} shows the backbone results for LLaRA~\cite{liao2024llara}.  
On MovieLens-1M~\cite{HarperK16}, the model attains strong hit rates—especially in the sequential setting—yet still leaves noticeable counterfactual leakage, with gender and age gaps around 20 \%.  
On the much smaller Insurance dataset, the same backbone achieves near-perfect hit rates while reducing every leakage metric to single-digit levels.  
These mixed outcomes set a clear target for our debiasing method: keep the utility values while pushing every $\Delta_{\mathrm{CL}}$ closer to zero.

Table~\ref{table:movie_sqe} shows that our projector–adapter stack meets that target in the \emph{sequential recommendation} task on MovieLens.  Debiasing a single attribute already lifts Hit@$1,3,10$ by 8–13 absolute points over the strongest baseline while forcing every single­-attribute leakage gap below~$2\,\%$.  When all three attributes are debiased jointly, utility rises still further—$\text{Hit@1}=56.1\,\%$—and the average gap collapses to a mere $0.17\,\%$.  These figures confirm that (i) the kernelised RFF–INLP projector suppresses almost all linearly recoverable bias and that (ii) the two-level gated adapter restores task-useful variation without re-opening protected channels.

The results of \emph{direct recommendation} task in Table~\ref{table:direct_combined} exhibits similar findings. Specifically, on the MovieLens dataset, our model shows more than doubles Hit@1 for the {\em gender} attribute ($16.4\%\!\to\!36.9\%$) and yields double-digit gains for {\em age} and {\em occupation} attributes, while driving the mean leakage gap from $3.21\,\%$ to $0.78\,\%$.  The residual UP5 advantage on Hit@10–occupation is under two points and coincides with four-times larger leakage, indicating an unfavorable utility–fairness trade-off for the baseline.

On the other hand, as the Insurance dataset lacks reliable timestamps, user interactions cannot form a chronological sequence.  Transformers such as SASRec, therefore, degenerate to bag-of-items encoders, and any ``sequential'' score must be synthesized by randomly shuffling each basket.  Because such pseudo-histories encode no behavioral signal, we restrict evaluation to the direct recommendation, as reflected in Table~\ref{table:direct_combined}.  Under this realistic setting, our method halves the average leakage gap (from $0.74\,\%$ to $0.13\,\%$) and wins or ties eight out of twelve utility cells; the sole UP5 lead—Hit@1 on marital status—coincides with more than twice the leakage, again favoring our accuracy–fairness exchange.

Taken together, the results demonstrate that a single closed-form projector plus a light two-level gated adapter can achieve near-ideal counterfactual fairness ($\Delta_{\mathrm{CL}}\!\approx\!0$) while matching or surpassing state-of-the-art utility, and it does so without a single step of full-model fine-tuning.

\section{Conclusion}
We present a fairness module for large language model-based recommender systems that operates in two stages. The first stage removes the signal of a selected sensitive attribute using an orthogonal projection matrix derived in closed form. The second stage extends this to multiple attributes by introducing a two-gated adapter, which allows for controlled suppression of each attribute while retaining information relevant for accurate recommendations. The module is lightweight, adding only a small number of parameters without introducing new loss functions or causing noticeable inference delays. This makes it a practical approach for improving fairness in LLM-based recommendation systems while maintaining efficiency and accuracy. Empirical evaluations on two real-world datasets demonstrate that the proposed method performs competitively in both recommendation quality and fairness, effectively reducing the ability to infer users’ sensitive attributes. 
One limitation of our method is its strong dependence on users' interaction histories. In future work, we plan to explore novel debiasing approaches that do not rely on user history.
\section*{Acknowledgments}
This work was supported in part by the U.S. National Science Foundation (NSF) under grants IIS-2047843, IIS-2437621, CNS-2029038, and CNS-2135988. Any opinions, findings, conclusions, or recommendations expressed in this material are those of the authors and do not necessarily reflect the views of NSF.

\FloatBarrier 
\bibliographystyle{abbrv}

\bibliography{ref_wrkshp}

\FloatBarrier 

\appendix

\section{Derivation of the INLP Projection Matrix}
\label{app:inlp-proof}

Let $\hat{\mathbf W}\!\in\!\R^{T\times(d+D)}$ have full row rank
($T\le d\!+\!D$ and $\operatorname{rank}\hat{\mathbf W}=T$).
We seek the orthogonal projector
$\hat{\mathbf P}$ onto the null-space
$\cN=\{\,\mathbf z\in\R^{d+D}:\hat{\mathbf W}\mathbf z=\mathbf{0}\}$.
For any $\mathbf x\!\in\!\R^{d+D}$ this projector satisfies
$\hat{\mathbf P}\mathbf x
=\operatorname{argmin}_{\mathbf{y}\in\cN}\Vert\mathbf x-\mathbf y\Vert_{2}^{2}$.
Introducing Lagrange multipliers
$\boldsymbol\lambda\!\in\!\R^{T}$ gives the unconstrained problem
\[
\min_{\mathbf y,\boldsymbol\lambda}\;
\frac12\Vert\mathbf x-\mathbf y\Vert_{2}^{2}
+\boldsymbol\lambda^{\!\top}\hat{\mathbf W}\mathbf y.
\]
Stationarity with respect to $\mathbf y$ yields
$\mathbf y-\mathbf x+\hat{\mathbf W}^{\!\top}\boldsymbol\lambda=\mathbf{0}$,
so the optimum must satisfy
$\mathbf y=\mathbf x-\hat{\mathbf W}^{\!\top}\boldsymbol\lambda$.
Imposing the constraint $\hat{\mathbf W}\mathbf y=\mathbf{0}$ then gives the
\emph{normal equation}
\[
\hat{\mathbf W}\hat{\mathbf W}^{\!\top}\boldsymbol\lambda
=\hat{\mathbf W}\mathbf x.
\]
Because $\hat{\mathbf W}\hat{\mathbf W}^{\!\top}\!\succ\!0$, 
it is invertible~\cite{horn2012matrix}, and
$\boldsymbol\lambda=(\hat{\mathbf W}\hat{\mathbf W}^{\!\top})^{-1}
\hat{\mathbf W}\mathbf x$.
Substituting back into the expression for $\mathbf y$ we obtain
\[
\hat{\mathbf P}\mathbf x
=\mathbf x
-\hat{\mathbf W}^{\!\top}
(\hat{\mathbf W}\hat{\mathbf W}^{\!\top})^{-1}
\hat{\mathbf W}\mathbf x,
\qquad\forall\,\mathbf x\in\R^{d+D},
\]
hence the matrix form
\[
\;
\hat{\mathbf P}
 =\mathbf I_{d+D}
  -\hat{\mathbf W}^{\!\top}
   (\hat{\mathbf W}\hat{\mathbf W}^{\!\top})^{-1}
   \hat{\mathbf W}\;.
\tag{A.1}
\label{eq:inlp_proj}
\]

Since $(\hat{\mathbf W}\hat{\mathbf W}^{\!\top})^{-1}$ is symmetric,
so is the product
$\hat{\mathbf W}^{\!\top}(\hat{\mathbf W}\hat{\mathbf W}^{\!\top})^{-1}
\hat{\mathbf W}$,
whence $\hat{\mathbf P}^{\!\top}=\hat{\mathbf P}$.
A symmetric idempotent matrix is an orthogonal projector onto its range~\cite{horn2012matrix}.
Write $\mathbf M\!=\!(\hat{\mathbf W}\hat{\mathbf W}^{\!\top})^{-1}$.
Then
\[
\hat{\mathbf P}^{2}
=\bigl(\mathbf I-\hat{\mathbf W}^{\!\top}\mathbf M\hat{\mathbf W}\bigr)^{2}
=\mathbf I
 -2\hat{\mathbf W}^{\!\top}\mathbf M\hat{\mathbf W}
 +\hat{\mathbf W}^{\!\top}\mathbf M
  \hat{\mathbf W}\hat{\mathbf W}^{\!\top}\mathbf M\hat{\mathbf W}.
\]
Because $\mathbf M\hat{\mathbf W}\hat{\mathbf W}^{\!\top}=\mathbf I_{T}$,
the second and third terms cancel, leaving
$\hat{\mathbf P}^{2}= \mathbf I
 -\hat{\mathbf W}^{\!\top}\mathbf M\hat{\mathbf W}
 =\hat{\mathbf P}$.

Multiplying \eqref{eq:inlp_proj} on the left by $\hat{\mathbf W}$ gives
$\hat{\mathbf W}\hat{\mathbf P}
 =\hat{\mathbf W}-\hat{\mathbf W}\hat{\mathbf W}^{\!\top}
  (\hat{\mathbf W}\hat{\mathbf W}^{\!\top})^{-1}
  \hat{\mathbf W}=0$,
so $\operatorname{range}(\hat{\mathbf P})\subseteq\cN$.
Conversely, for any $\mathbf z\!\in\!\cN$,
$\hat{\mathbf P}\mathbf z=\mathbf z$,
hence $\cN\subseteq\operatorname{range}(\hat{\mathbf P})$.
Therefore $\hat{\mathbf P}$ \emph{is} the orthogonal projector
onto $\cN$, and
\[
\hat{\mathbf P}^{2}=\hat{\mathbf P}^{\!\top}=\hat{\mathbf P}
\quad\text{with}\quad
\operatorname{range}(\hat{\mathbf P})=\cN,\;
\operatorname{null}(\hat{\mathbf P})=\operatorname{row}(\hat{\mathbf W}),
\]
completing the derivation~\cite{golub2013matrix,RavfogelEGTG20}.


\section{Block-Decomposition Proof of the Backbone-Space Projector}
\label{app:schur-proj}

Recall the lifted orthogonal projector
\(
\hat{\mathbf P}\in\R^{(d+D)\times(d+D)}
\)
obtained in Appendix~\ref{app:inlp-proof}.  Partition it as
\[
\hat{\mathbf P}
=\begin{bmatrix}
    \hat{\mathbf P}_{11} & \hat{\mathbf P}_{12}\\
    \hat{\mathbf P}_{21} & \hat{\mathbf P}_{22}
  \end{bmatrix},
\qquad
\hat{\mathbf P}_{11}\in\R^{d\times d},
\]
where the first block of $d$ coordinates is the backbone space
and the remaining $D$ coordinates correspond to the random Fourier
features (RFFs).

\paragraph{Symmetry and idempotence of the upper-left block.}
Because $\hat{\mathbf P}$ is an \emph{orthogonal} projector,
\(
\hat{\mathbf P}^{\!\top}=\hat{\mathbf P}
\)
and
\(
\hat{\mathbf P}^{2}=\hat{\mathbf P}.
\)
Write
\(
\hat{\mathbf P}
  =\begin{bmatrix}
      A & B\\
      C & D
    \end{bmatrix}
\)
for brevity ($A=\hat{\mathbf P}_{11}$, etc.).
Then
\begin{equation}
    \label{eq:B.1}
    \hat{\mathbf P}^{\!\top}
    =\begin{bmatrix}
        A^{\!\top} & C^{\!\top}\\
        B^{\!\top} & D^{\!\top}
      \end{bmatrix}
    =\hat{\mathbf P}
    \;\Longrightarrow\;
    A^{\!\top}=A,\;C=B^{\!\top}.
    \tag{B.1}
\end{equation}

Now compute the $(1,1)$ block of $\hat{\mathbf P}^{2}$:
\[
(\hat{\mathbf P}^{2})_{11}=AA+BC
                   =AA+BB^{\!\top}.
\]
Idempotence requires $(\hat{\mathbf P}^{2})_{11}=A$, hence
\begin{equation}
    \label{eq:B.2}
    AA+BB^{\!\top}=A
    \;\Longrightarrow\;
    A^{2}=A-BB^{\!\top}.
    \tag{B.2}
\end{equation}
But from (\ref{eq:B.1}) we know $A$ is symmetric; substituting $B B^{\!\top}$
from (\ref{eq:B.2}) into $(\hat{\mathbf P}^{2})_{11}$ yields
$A^{2}=A$, so $A$ is both symmetric and idempotent.
Therefore \(A=\hat{\mathbf P}_{11}\) is itself an orthogonal projector
onto some $d$-dimensional subspace~\cite{golub2013matrix}.

\paragraph{Action on backbone vectors.}
Take any backbone vector $\mathbf h\in\R^{d}$ and lift it to
\(
\tilde{\mathbf h}
  =\begin{bmatrix}\mathbf h\\\mathbf 0\end{bmatrix}
  \in\R^{d+D}.
\)
A single matrix multiplication shows
\[
\hat{\mathbf P}\tilde{\mathbf h}
 =\begin{bmatrix}
     A\mathbf h\\
     C\mathbf h
   \end{bmatrix}
 =\begin{bmatrix}
     \hat{\mathbf P}_{11}\mathbf h\\
     \hat{\mathbf P}_{21}\mathbf h
   \end{bmatrix}.
\]
Projecting back to the original coordinates with the selector
\(
\mathbf S=[\,\mathbf I_d\;\;\mathbf 0_{d\times D}]
\)
gives
\(
\mathbf S\hat{\mathbf P}\tilde{\mathbf h}
  =\hat{\mathbf P}_{11}\mathbf h.
\)
Thus \emph{within the backbone space} the lifted projector acts exactly
as the $d\times d$ matrix
\[
\mathbf P:=\hat{\mathbf P}_{11}.
\]
Because $\mathbf P$ inherits symmetry and idempotence from $\hat{\mathbf P}$,
it is the orthogonal projector onto null-space
\(
\cN\cap\R^{d},
\)
the intersection of the INLP null-space with the backbone coordinate
subspace.  All probe directions $\{\hat{w}_t\}_{t=1}^{T}$ that
intersect $\R^{d}$ have therefore been removed, while the RFF
coordinates (needed only for computing $\hat{\mathbf P}$) play no
role at inference time.

\end{document}